# Detection of developmental language disorder in Cypriot Greek children using a neural network algorithm

Georgios P. Georgiou[1] & Elena Theodorou[2]

[1]Department of Languages and Literature, University of Nicosia, Nicosia, Cyprus

[2]Department of Rehabilitation Sciences, Cyprus University of Technology, Limassol, Cyprus

**Abstract**

Children with developmental language disorder (DLD) encounter difficulties in acquiring various language structures. Early identification and intervention are crucial to prevent negative long-term outcomes impacting the academic, social, and emotional development of children. The study aims to develop an automated method for the identification of DLD using artificial intelligence, specifically a neural network machine learning algorithm. This protocol is applied for the first time in a Cypriot Greek child population with DLD. The neural network model was trained using perceptual and production data elicited from 15 children with DLD and 15 healthy controls in the age range of 7;10 – 10;4. The k-fold technique was used to crossvalidate the algorithm. The performance of the model was evaluated using metrics such as accuracy, precision, recall, F1 score, and ROC/AUC curve to assess its ability to make accurate predictions on a set of unseen data. The results demonstrated high classification values for all metrics, indicating the high accuracy of the neural model in classifying children with DLD. Additionally, the variable importance analysis revealed that the language production skills of children had a more significant impact on the performance of the model compared to perception skills. Machine learning paradigms provide effective discrimination between children with DLD and those with TD, with the potential to enhance clinical assessment and facilitate earlier and more efficient detection of the disorder.

**Introduction**

Developmental Language Disorder (DLD) is a neurodevelopmental communication-affecting disorder impacting the language processing regions of the human brain (Lee et al., 2020). This condition has a notable impact on a considerable percentage of the global child population, estimated at 7,58% (Norbury et al., 2016). Children with DLD experience difficulties in the perception and production of language structures as shown by a great number of studies. These may include phonology (Aguilar-Mediavilla et al., 2002; Georgiou & Theororou, 2023a), morphology (Calter et al., 2023; Marchman et al., 1999), syntax (Georgiou & Theodorou, 2023b; Novogrodsky & Friedmann, 2006), semantics (Gladfelter et al., 2019; Haebig et al., 2017), and pragmatics (Adams, 2002; Osman et al., 2011). Performance in these language domains can be used as a marker of DLD as it often differentiates between children with DLD and children with typical development (TD). DLD is not solely characterized by linguistic challenges, but it also encompasses behavioral, emotional, and social difficulties that may persist into adulthood (Clair et al., 2011). It is thus important to detect the disorder as soon as possible and apply effective interventions by speech language pathologists.

Clinical diagnosis of DLD seldom occurs before school age, despite reports from parents indicating delays in their children's early communication milestones from infancy (Borovsky et al., 2021). Even when it occurs, it is not an easy procedure. As a consequence, children are often underdiagnosed, meaning they are not identified as having DLD or misdiagnosed with another disorder such as attention deficit hyperactivity (Graham & Tancredi, 2019; Grimm & Schulz, 2014). Diagnostic inconsistencies can stem from either the absence of the area of children's difficulty in the assessment tool or the inability of the tool to measure the linguistic competence of children (Schwartz, 2009). There is a necessity to establish optimized tools, which can provide fast and reliable identifications. With the evolution of technology, artificial intelligence has been widely used for the detection of DLD. Different machine learning algorithms trained with language features successfully identified DLD with high predictive accuracy (Beccaluva et al., 2023; Gabani et al., 2014; Kotarba, & Kotarba, 2020; Oliva et al., 2014; Sharma, & Singh, 2022, Reddy et al., 2020). Nevertheless, the majority of previous work, including the aforementioned studies, trained the algorithms using acoustic features or features elicited from the production of language structures. Fewer studies have concentrated on the perception of language structures and other production measures aside from acoustic information.

This study aims to develop an automated approach capable of differentiating children with DLD from children with TD based on various perception and production language measures. Specifically, we utilize a neural network machine learning algorithm which has better prediction capabilities in linguistic data compared to more traditional models (for a comparison with linear discriminant analysis and C5.0, see Georgiou, 2023). The artificial neural network stands as a potent machine learning algorithm, drawing inspiration from the intricacies of the human brain (Georgiou, 2023). Comprising approximately 85 million neurons (Lantz, 2015), the human brain engages in intricate signal transmission: environmental stimuli or upstream neurons transmit signals to neuron dendrites, where signal processing occurs in the cell body before being conveyed along the axon to reach the output terminal (Zhang, 2016). In mirroring this biological process, the neural network processes input signals, representing feature variables crucial for pattern recognition. Each variable is assigned a weight corresponding to its relative importance. The activation function, akin to the human brain's signal processing, sums and processes these assigned signals. The network is organized into layers: the input layer, hidden layers, and output layer (Haykin, 2008). The input layer acts as the initial point, receiving raw input data. Positioned between the input and output layers, the hidden layers progressively transform input data into a more meaningful representation through intricate computations and learned patterns. Ultimately, the output layer generates the predicted output based on the transformed input from the hidden layers.

We provide metrics on the performance of the neural network model using data from Cypriot Greek children. The algorithm was trained with the responses of children elicited from language perception tests (voicing discrimination, syntactic and semantic grammaticality judgment tests) as well as their response time in these tests, and from their scores in various speech production tests

(production of vocabulary and morphosyntax, and sentence repetition). The selection of these language features is informed by previous research on Cypriot Greek and other populations with DLD, which has consistently highlighted challenges in these specific domains. Notably, Georgiou and Theodorou (2023a) observed that Cypriot Greek children with DLD demonstrated poorer performance in phonological and grammatical perception compared to their typically developing peers. Similarly, studies on Cypriot Greek children have shown difficulties in morphosyntactic production and sentence repetition (Petinou et al., 2023; Theodorou & Grohmann, 2015; Theodorou et al., 2017). In terms of reaction times, findings from international literature present a mixed picture. For instance, Zapparata et al. (2023) conducted a meta-analysis revealing varied reaction times among children with DLD across different tasks, stimulus types, modalities, and response modes in comparison to age-matched neurotypical groups. Conversely, Rakhlin et al. (2014) reported no significant differences in reaction times between Russian-speaking children with DLD and typically developing peers in a task assessing grammatical gender and agreement.

In this study, we employed a feed-forward neural network with a single hidden layer for the generation of predictions. Predictive machine learning paradigms have been used with success for the detection of DLD in the past in populations with different native languages (e.g., for Italian, see Beccaluva et al., 2023, for Czech, see Kotarba, & Kotarba, 2020; for Spanish, see Oliva et al., 2014) and are introduced for the first time in a Cypriot Greek population with DLD. Such a venture is important for several reasons. First, the ability of the machine learning algorithm to learn the language features of DLD and provide early and accurate identification of the disorder is of utmost importance as timely interventions can take place, leading to more favorable outcomes for this population of children. Second, it will substantially decrease both the cost of treatment and the workload of speech-language pathologists, allowing them to dedicate more time to enhancing the quality of their services.

**Methodology**

**Participants**

Thirty children participated in the study. Fifteen of them were children with DLD ($n_{females}$ = 7) in the age range of 7;10 – 10;4 ($M_{age}$ = 8;9, $SD$ = 1;5). These children were initially diagnosed by speech-language pathologists based on their clinical judgements. Another group of 15 children with TD ($n_{females}$ = 7) in the age range of 7;10 – 10;0 ($M_{age}$ = 8;11, $SD$ = 1;2) participated in the study. All children completed the Diagnostic Verbal IQ (DVIQ) test (Stavrakaki & Tsimpli, 2000), which provides measures of their vocabulary production, morphosyntactic production, metalinguistic abilities, morphosyntactic comprehension, and sentence repetition skills. In this study, we focus just on three measures concerning their language production skills, which were found critical for distinguishing Cypriot Greek children with DLD (e.g., see Petinou et al., 2023; Theodorou et al., 2017). We used the Greek version of the DVIQ test adjusted for the Cypriot Greek population. The assessment was framed in the form of a game. Children were tasked with naming pictures and repeating sentences. For example, in the vocabulary production subtest, children were presented with a picture of children engaged in play, and they were asked the question "what are the children doing?" ("they are playing"). In the morphosyntax test, the objective was to elicit the production of various morphosyntactic structures. For instance, children were shown pictures illustrating a cat on a bed, and researchers prompted them with a question like "This is a cat. Where is the cat?" ("The cat is on the bed"). In addition, the participants were exposed to spoken sentences, such as "I haven't studied my lessons". After hearing the sentence, the children were asked to repeat it exactly as they heard it. The children's responses from the DVIQ test were documented on the test answer sheets, subsequently scored, and analyzed at a later stage by the research team. Each correct response received 1 point, except for the sentence repetition subtest which was scored based on the number of errors in each repetition (maximum score of 3 points correct for each sentence). In addition, all children completed the Raven's Colored Progressive Matrices test (Raven et al., 2003) to estimate their nonverbal intelligence (IQ). Both groups of children with DLD and children with TD did not differ in terms of chronological age [$t(28) = –0.46, p = 0.64$] and nonverbal IQ [$t(28) = –1.27, p = 0.21$]. The following inclusion and exclusion criteria were applied for the selection of children: a) they should be native speakers of Cypriot Greek, b) they should permanently live in Cyprus, c) they should not have been diagnosed with neurological, cognitive, or mental impairment, and d) they should have normal hearing and vision as reported by their parents/caregivers and speech-language pathologists.

**Materials**

As part of the perceptual task, there were three distinct tests: one focused on phonology, another on grammar, and a third on semantics. The phonology test consisted of five Cypriot Greek fricative/stop voiced consonants [b d g v z] and their voiceless counterparts [p t k f s] embedded in a trisyllabic /CCV.'CV.CV/ (C=consonant, V=vowel) context, where the target consonant was at the initial position of the consonantal cluster; the latter context poses difficulties for children with DLD (Talli, 2010). Words corresponded to the phonotactics of real words. For the grammar test, there were three distinct subtests: (a) a subject-verb agreement subtest, (b) a clitics subtest, and (c) a pu-relative clause subtest, each comprising 10 trials. The study materials were created by the authors and were provided in Cypriot Greek. The subject-verb agreement subtest consisted of five sentences with correct subject-verb agreement and five sentences with incorrect subject-verb agreement, e.g., "Κάθε μέρα, ο σκύλος *παίζει/παίζουν\** στον κήπο" ("Every day, the dog *plays/play\** in the garden"). The clitics subtest featured five sentences with correct clitic positioning and five sentences with incorrect clitic positioning e.g., "Κάθε μέρα, περιμένουν να *το φέρει/φέρει το\**" ("Every day, they wait to *bring it/it bring\**"). The pu-relative clauses subtest included five sentences that correctly used the pu-relative pronoun (similar to the relative pronouns "that" and "who" of English) and five sentences that incorrectly omitted the pu-relative pronoun e.g., "Κάθε μέρα, παίζει με τον Γιώργο, *που/Ø\** εν φίλος της/φίλος της" ("Every day, s/he plays with George, *who /Ø\** is her/his friend"). In the semantic test, there were 20 trials: 5 sentences with correct and 5 sentences with anomalous verbs, e.g., "Κάθε μέρα, το τηλέφωνο *χτυπά/τρέχει\** για πολλή ώρα" ("Every day, the phone *rings/runs\** for a long time"), and 5 sentences with correct and 5 sentences with anomalous nouns, e.g., "Κάθε μέρα, η Ελένη διαβάζει *την εφημερίδα/την εικόνα*" ("Every day, Helen reads the *newspaper/image\**"). The materials were digitally audio recorded by one adult Cypriot Greek male speaker in a quiet room. The speaker was asked to read the material on paper maintaining a normal speaking rate. His productions were recorded using a professional audio recorder at a 44.1kHz sampling rate and saved as wav. files with a resolution of 24 bits. The output was adjusted for peak intensity in Praat (Boersma & Weenink, 2022).

## Procedure

The perceptual tests were scripted in Praat. In the phonology test, children participated in an AX discrimination test, which included both "different" trials (AB or BA) and "same" trials (AA or BB). They were seated in front of a computer monitor and asked to listen to pairs of words. Subsequently, they had to determine whether the words were acoustically the same or different by selecting either the "same" or "different" script labels. Each child completed a total of 20 trials, presented in a random order, which included 10 consonant pairs with two configurations each. In the grammar test, children were seated in front of a computer monitor and tasked with judging whether sentences played through the computer loudspeakers were syntactically correct or not. They indicated their judgments by clicking on the "correct" or "incorrect" labels. Each child evaluated 30 sentences. The semantic test followed a similar protocol to the grammar test. However, in this test, children were asked to identify whether different sentences were semantically correct or not. Each child evaluated 20 sentences. Throughout all the tests, stimuli were presented in a random order, with a brief optional two-minute break provided at the midpoint. Children wore headphones during all tests. An interstimulus interval of 700 ms was maintained (following Georgiou, 2021). The experiment allowed participants to replay the stimuli once by clicking on the "repeat" button. Prior to the main experiment, a familiarization test was conducted, which included four items for each test to ensure children understood the test requirements. No feedback was provided to the children.

## Training of the algorithm

The dataset was separated into two subsets: the training subset, which included 80% of the data and the testing subset, which included 20% of the data. A k-fold crossvalidation setup was established to assess the model's generalization performance. Specifically, 10 folds were utilized for crossvalidation. A systematic search for optimal hyperparameters was conducted to fine-tune the neural network model. The analysis explored various combinations of two hyperparameters: size (i.e., number of nodes) and decay, which is a regularization technique aiming to prevent overfitting (i.e., the phenomenon where the model becomes overly tailored to the nuances of its training data, thereby compromising its ability to generalize effectively to unseen data). The train function was applied to execute this grid search, resulting in a selection of the most suitable size and decay values. The maxit parameter was set to control the maximum number of iterations during the training process. Specifically, it determines the number of times the entire training dataset is processed by the network. Maxit was configured at the value of 100. Moreover, the MaxNWts parameter was employed to set a limit on the total number of weights in the network. This serves the purpose of preventing the model from becoming overly complex, which could result in overfitting. MaxNWts was set at 500. The logistic sigmoid activation function was employed since it is widely used in binary classification. With the best hyperparameters identified, the neural network model was trained on the training subset. Table 1 presents the model's hyperparameters. The model was trained using a neural network paradigm (NNET) from the *nnet* package (Ripley & Venables, 2023) in R (R Core Team, 2023) with GROUP serving as the response variable. This variable was transformed into a factor with two levels, namely DLD and TD. A set of language predictor variables was also used. This included PERCEPTION, that is, the number of correct/incorrect responses of children in the perceptual task, and RTPERCEPTION, that is, their reaction times in the perceptual task measured in seconds. The production measures of the DVIQ test were also included in the model. More specifically, the model included VOCABULARY, namely, the children's raw scores in the vocabulary production subset, MORPHOSYNTAX, namely, their raw scores in the morphosyntactic production subtest, and finally REPETITION, that is, their raw scores in their ability to repeat sentences in the sentence repetition subset.

Table 1: Model's hyperparameters using 10 folds

| Maxit | MaxNWts | Size | Decay |
|-------|---------|------|-------|
| 100   | 500     | 8    | 0.001 |

The trained neural network model was evaluated by assessing its performance on the testing subset. Key performance metrics, including accuracy, precision, recall, F1-score, and the area under the receiver operating characteristic (ROC) curve (AUC), were computed. Accuracy measures the proportion of correctly classified instances among all the instances in the dataset, providing an overall assessment of the model's correctness. It is calculated as follows: (True Positives + True Negatives) / (True Positives + True Negatives + False Positives + False Negatives). Precision quantifies the proportion of true positive predictions out of all positive predictions made by the model, demonstrating how well the model avoids false positives. The following formula is used for the calculation of precision: True Positives / (True Positives + False Positives). Recall assesses the model's ability to correctly identify all positive instances. It measures the proportion of true positive predictions out of all actual positive instances. It is calculated as follows: True Positives / (True Positives + False Negatives). The F1-score is a harmonic mean of precision and recall. It provides a balanced evaluation of a model's performance, considering both false positives and false negatives. The F1-score is calculated using the formula: 2 * (Precision * Recall) / (Precision + Recall). The ROC curve is a graphical representation of a model's ability to distinguish between classes at different threshold settings. The AUC quantifies the overall performance of the model in terms of its ability to separate positive and negative instances. The AUC is calculated by measuring the area under the ROC curve. A perfect model would have an AUC of 1, while a random model would have an AUC of 0.5. An AUC value closer to 1 yields a strong model, while an AUC value close to 0.5 suggests that the model's performance is no better than random guessing.

Finally, a variable importance analysis (VIA) was performed using the *caret* package (Kuhn et al., 2023). The analysis calculated variable importance to determine the relative contribution of predictor variables to the model's overall performance. Variable importance was assessed to understand the relevance of individual predictors in classification. The approach utilized is derived from Gevrey et al. (2003), which employs combinations of the absolute weights' values.

**Results**

Overall, the results demonstrated high classification performance values. As shown in Table 2, the highest value was observed for Recall, followed by ROC/AUC curve, accuracy, F1, and precision. The best size for the trained model was 8 (with the decay of 0.001) and exhibited high accuracy and Kappa, namely 0.91 ($SD$ = 0.05), and 0.83 ($SD$ = 0.11) respectively. Table 3 presents the model's accuracy and Kappa for each size. The architecture of the neural network model is illustrated in Figure 1. Figure 2 shows the ROC curves and the AUC values of both the training and testing subsets.

Table 2: Classification performance metrics for the trained neural network model

| Metric | Accuracy | Precision | Recall | F1 | ROC/AUC |
|---|---|---|---|---|---|
| Value | 0.90 | 0.87 | 0.92 | 0.89 | 0.90 |

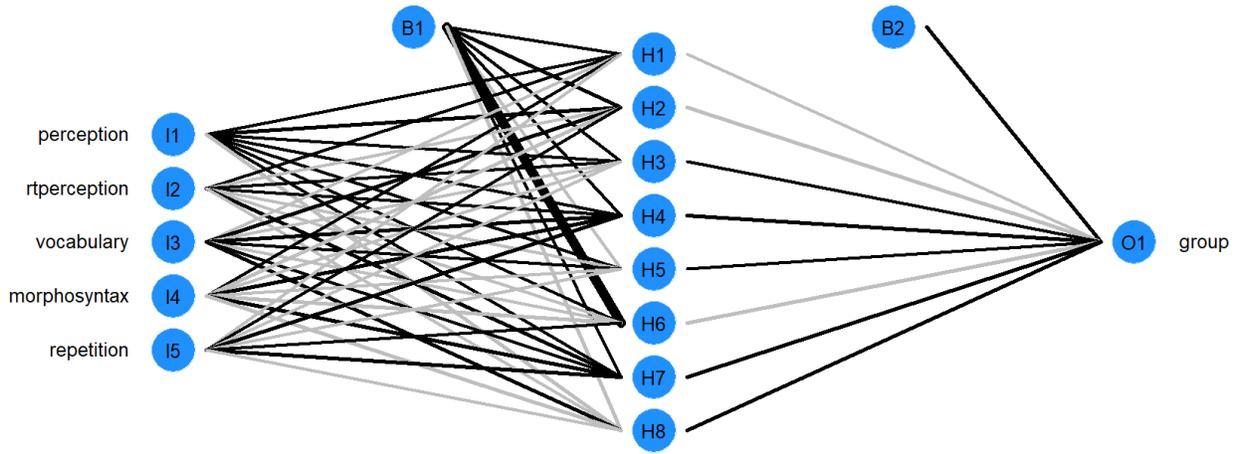

Figure 1: NNET architecture plot for the neural network model. I (predictor variables), B, H, and O (group: DLD and TD) represent the input layers, bias units, hidden layers, and output layers respectively. Each input is connected to every hidden neuron, which is represented by the lines, and the hidden neurons are further connected to the outputs.

Table 3: Accuracy and Kappa for each size of the trained model

| size | Accuracy | Accuracy SD | Kappa | Kappa SD |
|---|---|---|---|---|
| 1 | 0.573 | 0.156 | 0.146 | 0.312 |
| 2 | 0.690 | 0.204 | 0.381 | 0.408 |
| 3 | 0.795 | 0.126 | 0.589 | 0.253 |
| 4 | 0.819 | 0.176 | 0.638 | 0.352 |
| 5 | 0.793 | 0.158 | 0.587 | 0.316 |
| 6 | 0.851 | 0.137 | 0.702 | 0.274 |
| 7 | 0.907 | 0.056 | 0.814 | 0.112 |
| 8 | 0.914 | 0.053 | 0.829 | 0.107 |
| 9 | 0.909 | 0.055 | 0.818 | 0.110 |
| 10 | 0.879 | 0.136 | 0.758 | 0.273 |

*Tuning parameter decay was held constant at a value of 0.001

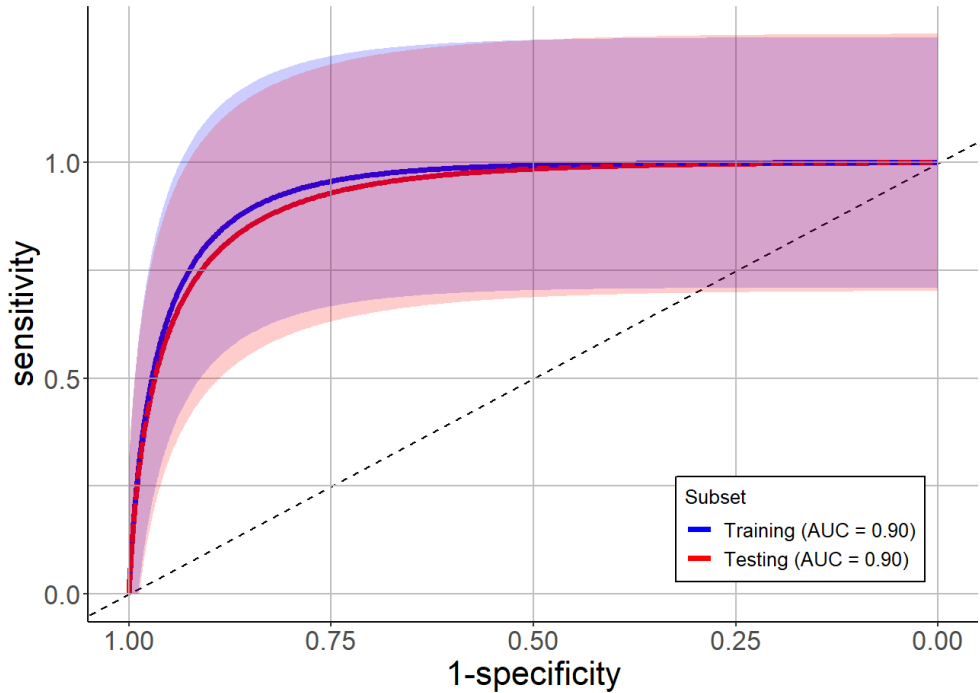

Figure 2: Mean ROC curves and AUC values of the training (blue) and testing (red) subsets. The dashed gray diagonal line shows the baseline. The shaded area indicated ± 1 standard deviation from the mean for two curves.

The next step included the implementation of VIA to identify the importance of each input feature in distinguishing children with DLD from children with TD. VIA showed that MORPHOSYNTAX was the most important variable (100), followed by VOCABULARY (86.4) and REPETITION (69.9). PERCEPTION (13.01) and RTPERCEPTION (0) were the least important variables. Figure 3 shows the importance of each variable based on the results of VIA.

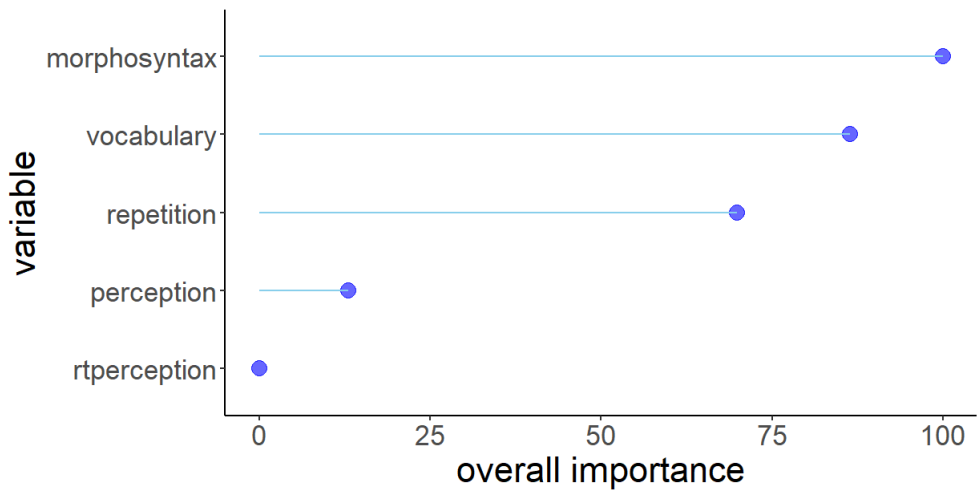

Figure 3: Overall importance of predictor variables indicated by VIA

**Discussion**

The aim of this study was to develop an automated method for the identification of DLD in Cypriot Greek children. To this end, we have trained a feedforward single-layer neural network on various language perception and production features collected from children with DLD and TD. The selection of features was based on the findings of previous studies that report challenges in children with DLD in the acquisition of these language structures. We provided an evaluation of various metrics to demonstrate the accuracy of the neural network in identifying children with DLD.

Our evaluation metrics indicated high overall performance (> 0.89) for the neural network model in detecting DLD. The evaluation process involved rigorous validation techniques to ensure the reliability of our results. Firstly, we utilized established methods such as k-fold cross-validation to assess the generalization performance of the neural network model. Furthermore, we conducted a grid search over a range of hyperparameters to optimize the model's architecture, including parameters such as the number of hidden units and the weight decay. Additionally, we employed well-established evaluation metrics such as accuracy, precision, recall, and the F1 score to quantify the model's performance. The dataset was split into two subsets, the training subset and the testing subset, which included 80% and 20% of the data correspondingly. This split is the most commonly used in neural networks (Ferentinos, 2018), offering optimal predictions (Seidu et al., 2022). The high values of metrics are observed on the testing subset suggesting that the model generalizes well to new, previously unseen data. Generalization is a key indicator of a model's ability to perform accurately in real-world scenarios beyond the training data. The fact that the model achieved great success on the testing subset implies that it has effectively captured underlying patterns in the data during training without memorizing specific instances, enhancing in that way its reliability and utility.

Another contribution of the study is the designation of the importance of each of the predictor variables in the model's classification performance. The most important variables were morphosyntactic production, vocabulary production, and sentence repetition. This corroborates a great number of earlier studies that highlight the importance of these features in indicating DLD (McGregor et al., 2013; Oetting et al., 2016; Theodorou et al., 2016; Zwitserlood et al., 2015). Responses in the perception tests and reaction times in these tests were the least important variables. While evidence suggests that language perception abilities differentiate between Cypriot Greek children with DLD and those with TD (e.g., Georgiou & Theodorou, 2023a), interestingly, this variable ranked second to last in importance. Furthermore, reaction times made no contribution to the model, aligning with studies that demonstrate no discernible difference in reaction times during language tasks between children with DLD and those with TD (e.g., Rakhlin et al., 2014). Therefore, the production abilities of children may be more informative about the identification of DLD compared to perception of language structures and reaction times.

An automated tool holds substantial promise as a valuable resource for clinicians in their endeavors to better predict DLD. Its predictive capabilities contribute to the potential contribution of artificial intelligence to clinical practice. Such a tool could serve as a complementary resource, working synergistically with traditional assessment methods for assessment and diagnosis. By leveraging technology to analyze linguistic patterns, the tool can provide timely and data-driven insights into children's linguistic trajectories. Enhanced diagnostic procedures among clinicians may lead to the early detection of DLD, thus enabling the implementation of timely intervention strategies. In addition, the cost of services will significantly reduce and clinicians will be able to enhance the quality of their work as they will save time through assessments. The integration of predictive models into clinical workflows holds the potential to revolutionize how language disorders are identified and addressed, ultimately improving outcomes for individuals with DLD.

However, the predictions provided by machine learning algorithms should be treated with caution. This is because the predictions depend on the quality of the input data and thus misleading outcomes can arise from erroneous, incomplete, or biased data (Toki et al., 2024). For example, imbalanced data can lead to problematic generalization capacity to new data for the algorithm (Song et al., 2022). Ultimately, configuring, interpreting, and validating models necessitate profound expertise and specialized skills, potentially restricting accessibility for certain researchers. Research endeavors aimed at devising models to distinguish individuals with particular conditions have underscored the intricate nature of model development and interpretation (Donelly et al., 2023). In addition, the lack of representativeness of groups based, for example, on gender and age can result in bias (Alam et al., 2022). Therefore, it is crucial for researchers to comprehend the functioning of these algorithms and the associated risks to ensure adherence to ethical standards and their reliable application in clinical settings.

## Conclusions

Although the algorithm yielded promising results, future endeavors should involve a larger participant pool to enhance the credibility of the findings. Moreover, future research can enrich the neural network algorithm with additional predictor variables such as acoustic information elicited from speech samples and measures of cognitive abilities as well as other variables such as sociolinguistic features, intervention history, etc. Considering the differences between children with DLD and children with TD in various linguistic and nonlinguistic aspects, the incorporation of these additional variables has the potential to refine the model's predictive capabilities. In this way, the algorithm may achieve heightened accuracy in distinguishing between these populations. This multifaceted approach could pave the way for more robust and tailored applications of artificial intelligence in the context of DLD assessment, offering valuable contributions to both research and practical diagnostic endeavors. Understanding the risks and challenges associated with the use of such tools in clinical practice is imperative to mitigate potential negative consequences.

## Declarations

### Funding

The study was funded by the "Metadidaktor" postdoctoral scheme, which was granted to the first author. The second author served as the scientific director of the project.

### Conflict of interests

The authors declare that they have no known competing financial interests or personal relationships that could have appeared to influence the work reported in this paper.

**Ethical approval**

The study has been approved by the Cyprus Bioethics Committee [ΕΕΒΚ/ΕΠ/2021/28]. Written consent for the participation of the children in the experiments was provided by both the children themselves and their parents or legal guardians, according to the declaration of Helsinki.

**Acknowledgments**

The study is supported by the Phonetic Lab of the University of Nicosia and the EPILOGO Lab of the Cyprus University of Technology. We would like to thank the child participants for their involvement in the study and their parents/caregivers for their consent and invaluable information, both of which were essential for the study's successful execution. We would also like to express our gratitude to the dedicated team of speech-language pathologists of the EPILOGO Lab, who played a vital role in data collection for this research project.